\documentclass{article}
\usepackage[main,final]{neurips_2025}
\usepackage[utf8]{inputenc}
\usepackage[T1]{fontenc}
\usepackage{hyperref}
\usepackage{url}
\usepackage{graphicx}
\usepackage{booktabs}
\usepackage{amsfonts}
\usepackage{nicefrac}
\usepackage{microtype}
\usepackage{xcolor}
\usepackage{enumitem}
\setitemize{leftmargin=*}
\usepackage{amsmath}
\usepackage{amssymb}
\usepackage{multirow}
\usepackage{tabularx}
\usepackage{array}
\usepackage{algorithm}
\usepackage{algpseudocode}
\usepackage{caption}
\usepackage{float}
\usepackage{bibunits}

\makeatletter
\renewcommand{\@notice}{}
\makeatother

\title{Multi-Scale Semantic Mapping in Urban Environments via \\ Observation Calibration and Policy Dependence Regularization}

\author{
Runling Long, Junhao Feng, Jia Wan \\
Harbin Institute of Technology, Shenzhen
}

\begin{document}

\maketitle

\begin{abstract}
Semantic mapping is fundamental to embodied navigation, yet existing methods are developed for indoor environments, where objects exhibit relatively limited scale variation and are observed from a restricted range of viewpoints. Urban environments pose substantially greater challenges: agents must map objects ranging from pedestrians to buildings while navigating large spaces with highly diverse viewing distances. These conditions introduce two key difficulties that existing datasets and methods fail to cover. First, object scale and observation distance can be severely mismatched. For example, small objects may be viewed from far away, whereas large objects may be observed at extremely close range, resulting in unreliable observation likelihoods. Second, objects with substantially different sizes and geometries require distinct mapping behaviors, which are difficult to capture with a single shared value estimator.
To investigate these challenges, we introduce a large-scale urban semantic mapping dataset featuring realistic city layouts, high-fidelity rendering, and instance-level annotations spanning multiple object scales. We then propose a category-aware likelihood calibration policy that identifies and alleviates unreliable observations according to object category and viewing distance. Because the calibration and motion policies are optimized toward the same mapping objective, they may learn redundant shortcuts and become excessively coupled. We therefore introduce a mutual-information (MI) regularizer that penalizes their estimated representation dependence and encourages complementary behaviors. To better model heterogeneous mapping strategies across object scales, we further employ category-wise value estimators. We formulate their joint optimization as a Pareto optimization problem to mitigate conflicting gradients across categories. Experiments demonstrate that our approach consistently outperforms state-of-the-art semantic mapping methods in challenging urban environments. The dataset and code will be publicly released.

\end{abstract}

\section{Introduction}

Semantic mapping transforms online visual observations into persistent spatial semantics for embodied AI. It has improved indoor navigation through semantic priors \cite{chaplot2020object, liang2021sscnav, majumdar2022zson, huang2023visual, zhou2023esc, gadre2023cows}, and is increasingly used as a grid- or graph-based representation in outdoor navigation \cite{yao2024aeroverse, gao2025openfly, wang2025uav, wang2024towards, liu2023aerialvln, ji2026towards, duan2026causalnav}. These trends make accurate semantic mapping essential for frontier navigation.

\begin{figure}[t]
    \centering
    \includegraphics[width=0.9\linewidth]{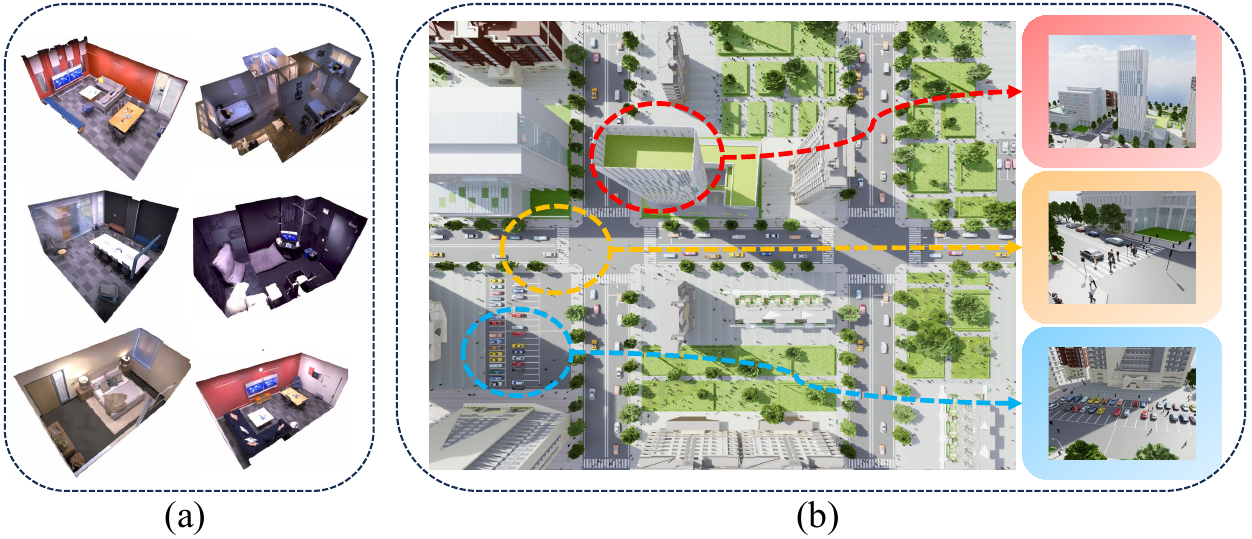}
    \caption{Existing indoor semantic mapping (a) vs. our multi-scale urban semantic mapping scenes (b). Ours explicitly considers the multi-scale objects and large navigable areas challenge, which is essential for outdoor navigation.}
    \label{fig:introduction}
\end{figure}

However, existing studies do not fully capture the challenges of semantic mapping in cities, where objects exhibit substantial scale variation. As illustrated in Fig.~\ref{fig:introduction}, existing indoor semantic mapping is conducted in compact spaces with nearby objects, while our task requires mapping multi-scale urban objects across larger navigable areas. We analyze this gap from two aspects.

From a dataset perspective, indoor navigation datasets \cite{chang2017matterport3d, ramakrishnan2021habitat, xia2018gibson} mainly contain household objects such as sofas, beds, and TVs, whose scales are relatively similar. Consequently, they cannot support multi-scale semantic mapping. Existing outdoor navigation datasets (e.g., \cite{ji2026towards, wang2024towards}) include multi-scale objects such as cars and buildings, yet they lack object-level annotations. This makes it impossible to extract semantic maps from these datasets, and thus makes them unsuitable for our task.

From a methodological standpoint, methods with the same objective as ours, i.e., actively constructing a semantic map, are mainly developed and evaluated indoors \cite{zhang2024active, chen2025understanding, chen2025activegamer, asgharivaskasi2023semantic, chen2024gennbv, chen2025GLEAM, li2025nextbestpath, siming2024active}. Although they can be applied to our task, their behavior in large-scale city scenes remains unclear, due to the limited variation in object scale and narrow navigation space of their datasets. Existing city-level navigation methods \cite{ji2026towards, duan2026causalnav} are typically designed for Vision-Language Navigation (VLN). Even when some methods use semantic maps, their task settings require the agent to only focus on limited objects in one episode. By contrast, our task requires full-scene semantic mapping. Therefore, these methods have substantially different task contexts from ours, and they lack specific designs for multi-scale objects.

Based on this gap analysis, we develop a simulator that explicitly reflects the multi-scale challenges in urban environments. The simulator covers object volumes from $0.01\,\text{m}^3$ to $208.47\mathrm{k}\,\text{m}^3$, including urban elements such as pedestrians, cars, and skyscrapers. The simulator uses Geographic Information System (GIS) \cite{chang2016geographic} to derive real-world street and block layouts, and then uses an LLM's common knowledge of urban environments to plan context-appropriate buildings and object distributions, ensuring authenticity and variety. A professional robotic simulator is then used to provide photo-level high-fidelity rendering with GPU parallelization. These designs model the visual conditions of real-world cities, and enable efficient data generation and agent training.

We then propose an RL agent that tackles the multi-scale challenges. Since the vision models used by the agent are not trained for each semantic-viewpoint distribution, they generate unreliable visual likelihoods when the viewpoint is suboptimal, e.g., observing pedestrians from far away while observing buildings from a very close range. To address this mismatch of object scale and viewing distance, we propose to train a likelihood calibration policy that estimates per-category map updating strength at grids to mitigate the effects of erroneous likelihoods. This module is trained along with the motion policy without fine-tuning vision models, improving mapping accuracy with a lightweight approach.

The calibration and motion policies are designed to play complementary roles.
However, because they are jointly optimized toward the same mapping objective,
their representations may become dependent through shared map-improving cues.
We mathematically illustrate that when such dependence arises, it can impair joint policy
optimization and lower performance. We therefore
introduce an MI-based regularizer that penalizes the estimated dependence
between their representations, alleviating the risk of redundant shortcut learning.

For policy optimization, multi-scale objects require different mapping policies because their optimal observation positions differ substantially. This makes it difficult for the original single value predictor to model policy advantages due to limited representation ability. To better model these advantages, we propose predicting values for each category. Since this may introduce gradient conflicts among different value estimators, we identify this as a Pareto optimization problem, and use a gradient balancing method to alleviate the conflicts. This modeling achieves the final performance improvement.

Our contributions are:

{
\vspace{0pt}
\begin{itemize}
    \setlength{\itemsep}{0pt}
    \setlength{\parsep}{0pt}
    \setlength{\topsep}{0pt}
    \setlength{\partopsep}{0pt}
    \item We formulate multi-scale semantic mapping in urban environments, highlighting the challenges introduced by extreme variations in object size and observation distance. To support research on this problem, we introduce a large-scale dataset with realistic city layouts, high-fidelity rendering, and multi-scale instance-level annotations.
    \item We propose a category-aware likelihood calibration policy that alleviates unreliable observations arising from mismatches between object scale and viewing distance. We further introduce an MI regularizer to encourage complementary behavior learning and restrict harmful dependence between the calibration and motion policies.
    \item We develop category-wise value estimators to capture the heterogeneous mapping strategies required by objects at different scales. To address gradient conflicts among these estimators, we formulate policy learning as a Pareto optimization problem that balances their objectives.
\end{itemize}
}

\begin{table*}[t]
\centering
{\small
\setlength{\tabcolsep}{0.8mm}
\begin{tabular*}{\textwidth}{@{\extracolsep{\fill}}l c c c c c c c@{}}
\toprule
\textbf{Dataset} & \textbf{Type} & \textbf{Platform} & \textbf{Scenes} & \textbf{Scale} & \shortstack{\textbf{Object volume}\\\textbf{($\text{m}^3$)}} & \shortstack{\textbf{Avg.}\\\textbf{objects/scene}} & \shortstack{\textbf{Obj.-level}\\\textbf{ann.}} \\
\midrule
OpenFly \cite{gao2025openfly} & VLN & UE4 & 21 & - & $10 - 100\mathrm{k}$ & - & $\times$ \\
EmbodiedCity \cite{gao2024embodiedcity} & VLN & UE5 & 1 & City & $2 - 100\mathrm{k}$ & - & $\times$ \\
\midrule
UrbanScene 3D \cite{lin2021urbanscene3d} & Map & UE4 & 16 & City & $10 - 100\mathrm{k}$ & 865.0 & $\times$ \\
GLEAM \cite{chen2025GLEAM} & Map & Habitat & 1152 & House & $0.2 - 5.0$ & - & $\times$ \\
MP3D \cite{chang2017matterport3d} & Sem-Map & Habitat & 90 & House & $0.2 - 5.0$ & 564.6 & $\checkmark$ \\
EmbodiedScan \cite{wang2024embodiedscan} & Sem-Map & Habitat & 5185 & House & $0.2 - 5.0$ & 30.9 & $\checkmark$ \\
\midrule
Ours & Sem-Map & Isaac Sim & 80 & City & $0.01 - 208.47\mathrm{k}$ & 9323.6 & $\checkmark$ \\
\bottomrule
\end{tabular*}
}
\caption{Dataset comparison. Ours contains objects spanning a wide scale range and explicit object-level annotations.}
\label{tab:dataset_comparison}
\end{table*}

\section{Related Work}

\subsection{Dataset}

Existing semantic mapping datasets are for indoor environments and cannot support multi-scale urban semantic mapping. Indoor datasets such as Replica \cite{straub2019replica}, ScanNet \cite{dai2017scannet}, and EmbodiedScan \cite{wang2024embodiedscan} provide 3D scans and semantic labels, but mainly contain household objects with limited scale variation and relatively small navigation areas. Outdoor datasets such as OpenFly \cite{gao2025openfly}, OpenUAV \cite{wang2024towards}, EmbodiedCity \cite{gao2024embodiedcity}, and UrbanScene3D \cite{lin2021urbanscene3d} contain city-level scenes and larger objects, but are often designed for VLN or navigation and lack object-level labels such as quantities, positions, and 3D meshes, making them unsuitable for our task.

Tab.~\ref{tab:dataset_comparison} compares these datasets. Our dataset explicitly measures object scales from $0.01\,\text{m}^3$ to $208.47\mathrm{k}\,\text{m}^3$ and provides object-level annotations for active mapping tasks.

\subsection{Active Semantic Mapping}

Active semantic mapping reconstructs a semantic map while planning viewpoints online. Existing full-map methods \cite{asgharivaskasi2023semantic, chen2025understanding, chen2025activegamer} often select views by map uncertainty over grid-based, NeRF, or 3DGS representations \cite{zhang2024active, siming2024active, li2025nextbestpath}. Learning-based methods \cite{chaplot2020learning, chen2024gennbv, chen2025GLEAM, guedon2023macarons} model future gains, but many focus on geometry or indoor scenes.

Other methods use semantic maps for Object Navigation \cite{georgakis2021learning, georgakis2022cross, ramrakhya2023pirlnav, georgakis2022uncertainty, zhang2024imagine}. These methods usually target one or a few objects rather than optimizing full-scene semantic reconstruction. Overall, existing approaches lack designs for large-scale outdoor perception and multi-scale planning; our method addresses this gap with likelihood calibration and scale-aware planning.

\begin{figure}[t]
    \centering
    \includegraphics[width=0.85\linewidth]{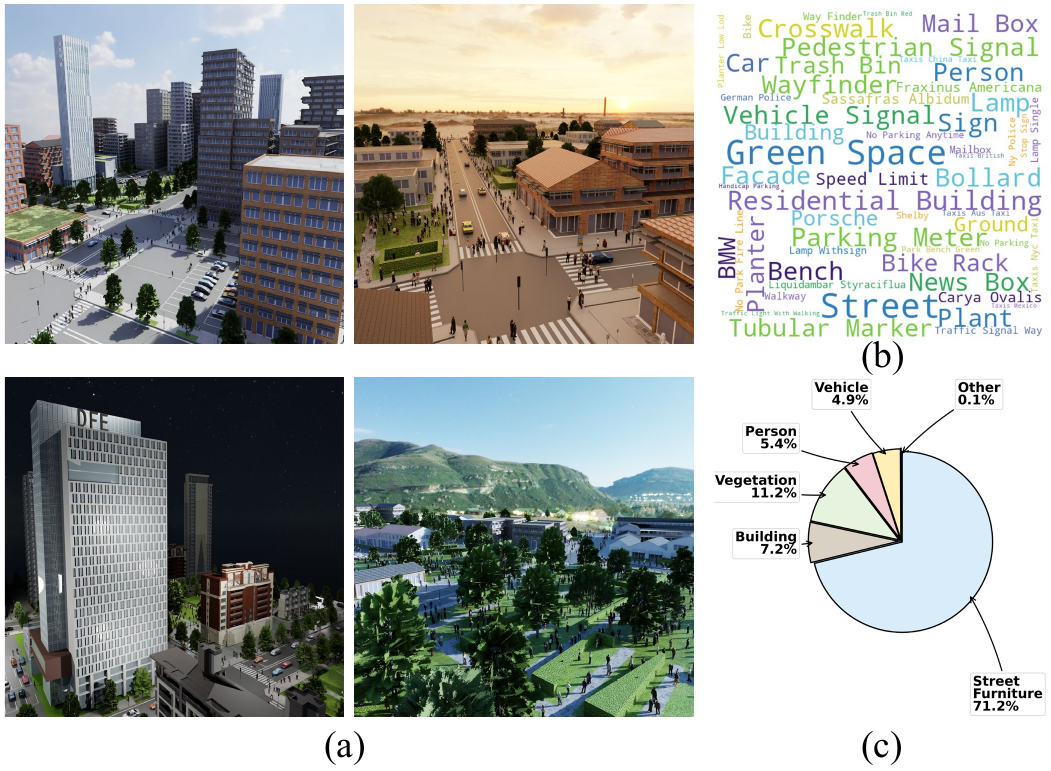}
    \caption{Dataset visualization. (a) High-fidelity visual conditions of the simulator. (b) Object name word cloud, showing the semantic diversity of generated urban objects. (c) Merged object category distribution.}
    \label{fig:dataset_new}
\end{figure}

\section{Dataset}

Existing outdoor navigation datasets lack object-level annotations, scene diversity, and control over multi-scale semantics. We therefore construct a fully simulated urban dataset by generating city structures and semantic object placements, then rendering annotated RGB-D observations under diverse visual conditions.

\subsection{Scene Generation}

We use CityEngine \cite{badwi20223d}, a professional city-planning tool widely used in the building industry, to plan city layouts. It imports real-world GIS data including street graph and building block layouts from georeferenced OSM street networks \cite{bennett2010openstreetmap}. We select diverse layouts covering real-world environments such as central business districts, towns, suburbs, and rural areas. The type of the GIS data is used for further planning.

We plan building block details with a hierarchical LLM-assisted process. Given the GIS scene type, the LLM uses its knowledge of real-world cities to assign block semantics such as residential, commercial, or public green areas. Conditioned on the scene and block types, it specifies crowd or vehicle distributions, architectural appearance, and visual style. This information is then input into CityEngine to generate assets. This process simulates urban spatial organization and object co-occurrence while enabling controlled scene diversity.

\subsection{Rendering for Robotic Training}

Isaac Sim RTX \cite{mittal2025isaac} renders the assets into RGB-D observations. It contains diverse lighting conditions such as sunny daytime, nighttime, and dusk. GPU parallel processing enables efficient robotic training across the simulated environments. Fig.~\ref{fig:dataset_new} summarizes their visual and semantic statistics, showing that our simulator provides high-fidelity data with diversity.

\section{Method}

We first define the task, then present three core components: likelihood calibration for unreliable observation alleviation, scale-calibration and motion dependence regularization for complementary behavior learning, and Pareto frontier exploration for balancing multi-scale value optimization. The overall framework is shown in Fig.~\ref{fig:framework}.

\subsection{Task Definition}

\begin{figure*}[t]
    \centering
    \includegraphics[width=0.8\textwidth]{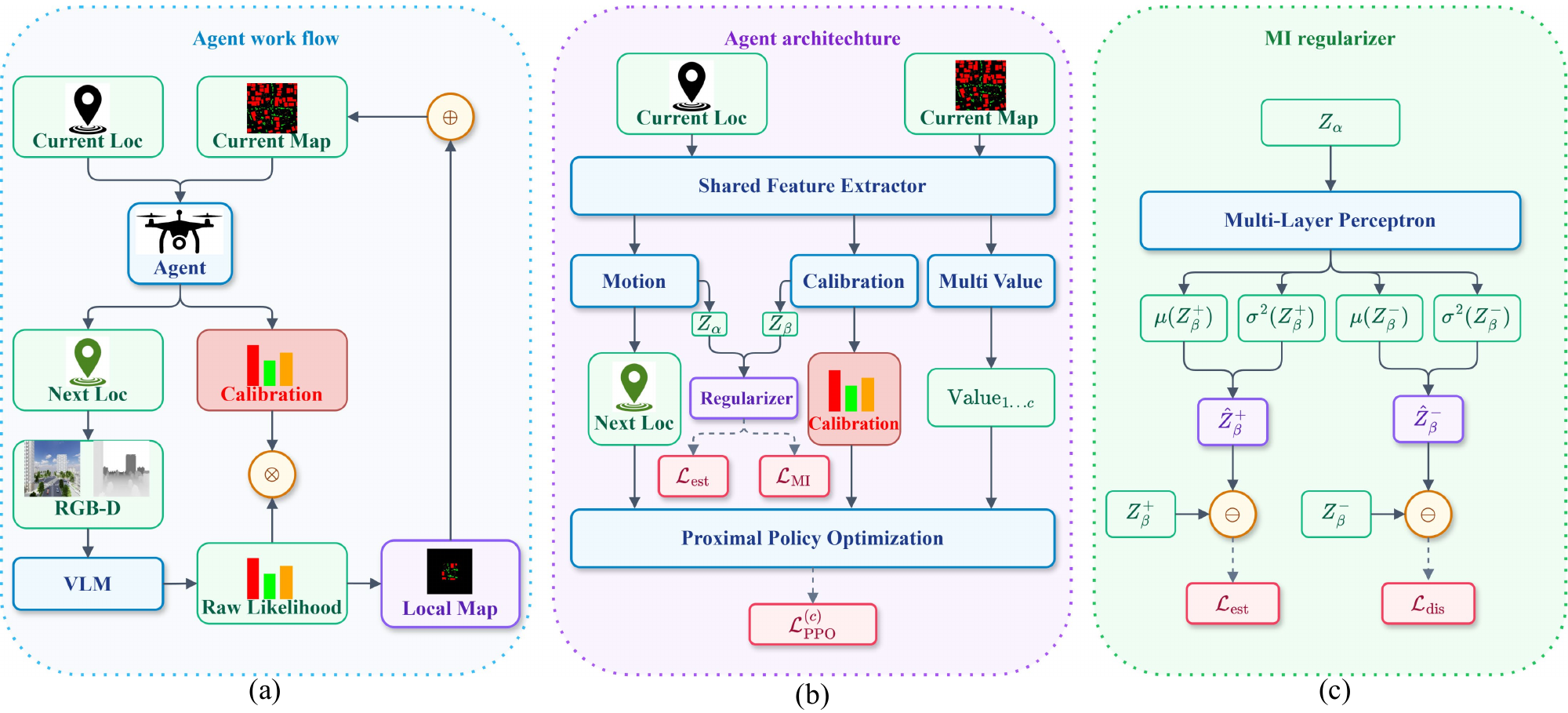}
    \caption{(a) The agent workflow. The agent additionally predicts a likelihood calibration. (b) The agent architecture, with a CLUB module for motion-perception dependence regularization and a multi-class value head. (c) The CLUB module uses the motion latent $Z_\alpha$ to estimate positive and negative perception latents $Z_\beta^+$ and $Z_\beta^-$. The resulting joint-marginal likelihood gap penalizes estimated MI during policy training.}
    \label{fig:framework}
\end{figure*}

In our task, an agent is initialized in an environment without any environmental priors. At each time step $i$, it captures RGB-D images, computes per-pixel semantic likelihoods using a VLM, and projects these egocentric likelihoods to a 2D plane to form a local semantic map $m_i$. At time step $t$, the agent's observation is defined as the historical context $z_{1:t}=\{(m_i,\mathbf{p}_i)\}_{i=1}^{t}$, where $\mathbf{p}_i$ denotes the historical agent pose. The current local map $m_t$ is fused into the global map, and the motion policy $\pi_{\theta}(a \mid z_{1:t})$ predicts the next action distribution. After selecting the most probable action, the agent moves to the next location and repeats this procedure until reaching the maximum number of steps. The final global map is used as the semantic reconstruction. This workflow is demonstrated in Fig.~\ref{fig:framework}(a).

\subsection{Likelihood Calibration}

In large-scale scenes, multi-scale objects are often observed from suboptimal positions due to large navigable spaces. Since the vision module is not trained for every semantic-spatial configuration, such observations may produce unreliable likelihoods (e.g., pedestrians at $100\,\text{m}$ vs. buildings at $10\,\text{m}$). We therefore predict category-wise update strengths to mitigate the effects of such erroneous likelihoods.

We apply a standard Bayesian updating framework \cite{zhang2024active} to fuse the local and global semantic maps. In basic Bayesian updating, the observation likelihood is directly used to compute the map posterior. While this approach suffices in settings without mismatch, we calibrate the Bayesian updating rule to mitigate erroneous likelihoods caused by suboptimal observations in large-scale scenes:
\begin{equation}
\label{eq:parameterization}
    P(c \mid z_{1:t}, v)
    = \frac{\exp\left( \beta_{t,v,c} \cdot l_{v,c} \right)}
    {\sum_{j=1}^C \exp\left( \beta_{t,v,j} \cdot l_{v,j} \right)}.
\end{equation}
where $l_v \in \mathbb{R}^C$ denotes the raw VLM logits at voxel $v$, and $\beta_{t,v,c}\sim\pi_{\theta}^{\beta}(\cdot\mid z_{1:t},v)$ is the spatial-semantic calibration vector. Since the input of the policy contains historical positions and updated map, the calibration also captures historical context and local cues, including occlusion, range, and height. We train it jointly with the motion policy instead of fine-tuning the VLM. The calibrated likelihood is then fused into the global semantic map using the binary log-odds rule:
\begin{equation}
\label{eq:updating}
    L_{v,c}^{t} = L_{v,c}^{t-1} + \log \left( \frac{P(c \mid z_{1:t}, v)}{1 - P(c \mid z_{1:t}, v)} \right).    
\end{equation}
By applying Eqs.~\ref{eq:parameterization} and~\ref{eq:updating} along the trajectory, the semantic map is constructed.

\subsection{Scale Calibration and Dependence Regularization}

The calibration and motion policies are optimized with the shared semantic mapping objective. Yet this joint optimization risks coupling their representations around map-improving cues. Once the coupling is severe, one policy may learn shortcuts that depend on the other policy, instead of learning robust complementary calibration and planning behaviors. Such shortcuts may reduce generalization, since a failure mode in one policy may propagate to the other.

To analyze the drawback that potential policy coupling may introduce, we consider a variational formulation of map reconstruction \cite{cheng2025asymmetric}. We assume a latent encoding process $q_{\phi}(Z_\alpha, Z_\beta \mid z_{1:t})$, where $Z_\alpha$ and $Z_\beta$ represent the extracted motion and calibration policy features. The Evidence Lower Bound (ELBO) of the mapping objective is formulated as:
\begin{equation}
\begin{aligned}
    \mathcal{L}_{\text{ELBO}} 
    &= \mathbb{E}_{q_{\phi}(Z_\alpha, Z_\beta \mid z_{1:t})} \left[ \log p(M \mid Z_\alpha, Z_\beta) \right] \\ 
    &- \underbrace{\mathcal{D}_{KL} \left( q_{\phi}(Z_\alpha, Z_\beta \mid z_{1:t}) \parallel p(Z_\alpha, Z_\beta) \right)}_{{D}_{KL}}.
\end{aligned}
\end{equation}
Assuming a factorized prior \( p(Z_\alpha, Z_\beta) = p(Z_\alpha) p(Z_\beta) \), we expand the KL divergence term as in Eq. \ref{eq:KL}. For brevity, let $q_{\alpha\beta} = q_{\phi}(Z_\alpha, Z_\beta \mid z_{1:t})$, $q_\alpha = q_{\phi}(Z_\alpha \mid z_{1:t})$, and $q_\beta = q_{\phi}(Z_\beta \mid z_{1:t})$,
\begin{equation}
\begin{aligned}
    \mathcal{D}_{KL}
    &= \iint q_{\alpha\beta} \, \log \frac{q_{\alpha\beta}}{p(Z_\alpha) p(Z_\beta)} \, dZ_\alpha \, dZ_\beta \\
    &= \underbrace{\iint q_{\alpha\beta} \, \log \frac{q_{\alpha\beta}}{q_\alpha q_\beta} \, dZ_\alpha \, dZ_\beta}_{I_q(Z_\alpha; Z_\beta \mid z_{1:t})} \\
    &+ \underbrace{\int q_\alpha\,\log\frac{q_\alpha}{p(Z_\alpha)}\,dZ_\alpha}_{\mathcal{D}_{KL}\left(q_\alpha\parallel p(Z_\alpha)\right)} + \underbrace{\int q_\beta\,\log\frac{q_\beta}{p(Z_\beta)}\,dZ_\beta}_{\mathcal{D}_{KL}\left(q_\beta\parallel p(Z_\beta)\right)}.
\end{aligned}
\label{eq:KL}
\end{equation}
The term $I_q(Z_\alpha;Z_\beta\mid z_{1:t})$ is the MI between the two policies and measures the degree of their dependence. It lowers the ELBO when the marginal KL terms are fixed, indicating that the distributional divergence between the reconstructed map and the real map may increase. This motivates us to penalize the MI during policy optimization. Let $\mathcal{J}_{\mathrm{map}}(\phi)$ denote the expected return under the semantic mapping reward. We formulate the constrained policy objective as
\begin{equation}
\max_{\phi}\ \mathcal{J}_{\mathrm{map}}(\phi)
\quad \text{s.t.}\quad
I_q(Z_\alpha;Z_\beta\mid z_{1:t}) \leq \delta,
\label{eq:mi_constraint}
\end{equation}
The Lagrangian of Eq.~\ref{eq:mi_constraint} is $\mathcal{J}(\phi,k)=\mathcal{J}_{\mathrm{map}}(\phi)-k[I_q(Z_\alpha;Z_\beta\mid z_{1:t})-\delta]$, where $k\geq0$. Since $k\delta$ is a constant, it can be omitted during optimization, yielding an MI regularization term weighted by $k$. Note that since this regularizer is weighted and does not impose the stronger assumption of statistical independence, it only suppresses extreme policy coupling. As a result, the beneficial coupling is not completely eliminated.

To estimate the intractable MI term $I_q(Z_\alpha; Z_\beta \mid z_{1:t})$, we employ the Contrastive Log-ratio Upper Bound (CLUB) \cite{cheng2020club}. We introduce a variational predictor $q_{\mu}(Z_{\beta} \mid Z_{\alpha})$, parameterized by a neural network $\mu$, to estimate the conditional density of the perceptual latent given the motion latent. For a training batch of size $N$, the predictor is trained by minimizing the Negative Log-Likelihood:
\begin{equation}
    \mathcal{L}_\text{estimator} = - \frac{1}{N} \sum_{i=1}^{N} \log q_{\mu}(Z_{\beta}^{(i)} \mid Z_{\alpha}^{(i)}).
\end{equation}
During the policy update, $\mu$ is fixed and \(q_{\phi}(Z_\alpha, Z_\beta \mid z_{1:t})\) is optimized with the estimated MI penalty. The MI regularization loss is defined as the difference between the log-likelihood of joint samples and the average log-likelihood of marginal samples:
\begin{equation}
\begin{aligned}
    \mathcal{L}_\text{MI}
    &= \mathbb{E}_{q_{\phi}(Z_\alpha, Z_\beta \mid z_{1:t})} \left[ \log q_{\mu}(Z_\beta \mid Z_\alpha) \right] \\
    &- \mathbb{E}_{q_{\phi}(Z_\alpha \mid z_{1:t}) q_{\phi}(Z_\beta \mid z_{1:t})} \left[ \log q_{\mu}(Z_\beta \mid Z_\alpha) \right].
\end{aligned}    
\end{equation}

\subsection{Pareto Frontier Exploration}

For RL methods such as Proximal Policy Optimization (PPO) \cite{schulman2017proximal}, a single value head is used to estimate advantages. However, in multi-scale scenarios, objects at different scales require distinct mapping strategies. The size divergence requires the agent to move to different spatial positions to align with their optimal viewpoints. In such cases, a single value head cannot adequately model this complexity. To better model the advantages, we use a separate value prediction for each category. Since the value estimators share the same input features but have different optimization directions, their gradients may conflict. We use Pareto optimization to balance these gradients.

We first separate the coverage reward into a class-wise formulation:
\begin{equation}
\label{eq:reward}
r_{t+1}^{\mathrm{CR}} = \mathrm{CR}_{t+1}^{(c)} - \mathrm{CR}_{t}^{(c)},
\end{equation}
where $\mathrm{CR}_{t}^{(c)}$ is the coverage ratio of class $c$ at time $t$. The multi-category loss is then formulated as:
\begin{equation}
\label{eq:class_loss}
\mathcal{L}_\text{c} = \mathcal{L}_\text{PPO}^{(c)} + \frac{k}{C}\mathcal{L}_\text{MI},
\end{equation}
where $\mathcal{L}_\text{PPO}^{(c)}$ uses the category-specific advantage, $k$ controls MI regularization, and $\mathcal{L}_\text{estimator}$ is used only to train the CLUB predictor. We use Nash-MTL \cite{navon2022multi} to balance the category gradients. Let $g_c=\nabla_{\Theta}\mathcal{L}_c(\Theta)$ and $G=[g_1,\ldots,g_C]$. The Nash weights and shared-parameter update are
\begin{equation}
    G^\top G\alpha^*=\frac{1}{\alpha^*},\quad
    \alpha^*\in\mathbb{R}_{++}^{C},\qquad
    \Theta\leftarrow\Theta-\eta G\alpha^*,
\end{equation}
where $1/\alpha^*$ is the element-wise reciprocal. This bargaining update reduces dominance by any single object scale.

\begin{table*}[t]
\centering
{\small
\setlength{\tabcolsep}{2.0pt}
\begin{tabular*}{\textwidth}{@{\extracolsep{\fill}}l c c c c c c@{}}
\toprule
\multirow{3}{*}{\textbf{Method}} & \multicolumn{6}{c}{\textbf{CCR (\%)} $\uparrow$} \\
\cmidrule(r){2-7}
& \multicolumn{2}{c}{$c_{\text{small}}$} & \multicolumn{2}{c}{$c_{\text{medium}}$} & \multicolumn{2}{c}{$c_{\text{large}}$} \\
\cmidrule(r){2-3} \cmidrule(r){4-5} \cmidrule(r){6-7}
& CLIP & DINOv3 & CLIP & DINOv3 & CLIP & DINOv3 \\
\midrule
Uncertainty & 61.7\ensuremath{\pm}0.3 & 62.9\ensuremath{\pm}1.9 & 74.7\ensuremath{\pm}3.4 & 78.5\ensuremath{\pm}2.1 & 95.5\ensuremath{\pm}2.1 & 96.2\ensuremath{\pm}1.3 \\
Zhang et al. & 52.5\ensuremath{\pm}1.5 & 54.7\ensuremath{\pm}4.8 & 75.3\ensuremath{\pm}4.2 & 79.7\ensuremath{\pm}3.3 & 96.5\ensuremath{\pm}0.7 & 97.9\ensuremath{\pm}0.7 \\
RayFronts & 28.5\ensuremath{\pm}5.1 & 27.8\ensuremath{\pm}4.4 & 59.8\ensuremath{\pm}8.6 & 62.6\ensuremath{\pm}6.9 & 91.9\ensuremath{\pm}2.2 & 92.2\ensuremath{\pm}3.5 \\
ActiveSGM & 57.6\ensuremath{\pm}4.8 & 60.2\ensuremath{\pm}6.6 & 93.8\ensuremath{\pm}2.4 & 95.6\ensuremath{\pm}2.2 & 95.7\ensuremath{\pm}2.4 & 97.4\ensuremath{\pm}1.5 \\
GLEAM & 81.1\ensuremath{\pm}2.1 & 82.8\ensuremath{\pm}0.9 & 91.5\ensuremath{\pm}1.1 & 93.7\ensuremath{\pm}3.4 & 97.4\ensuremath{\pm}1.6 & 98.7\ensuremath{\pm}0.4 \\
\midrule
\textbf{Ours} & \textbf{90.0\ensuremath{\pm}1.6} & \textbf{93.2\ensuremath{\pm}1.0} & \textbf{95.9\ensuremath{\pm}1.3} & \textbf{98.9\ensuremath{\pm}0.8} & \textbf{98.8\ensuremath{\pm}0.6} & \textbf{99.4\ensuremath{\pm}0.6} \\
\bottomrule
\end{tabular*}
\par\medskip
\begin{tabular*}{\textwidth}{@{\extracolsep{\fill}}l c c c c@{}}
\toprule
\multirow{2}{*}{\textbf{Method}} & \multicolumn{2}{c}{\textbf{OCR (\%)} $\uparrow$} & \multicolumn{2}{c}{\textbf{Var} $\downarrow$} \\
\cmidrule(r){2-3} \cmidrule(r){4-5}
& CLIP & DINOv3 & CLIP & DINOv3 \\
\midrule
Uncertainty & 77.3\ensuremath{\pm}0.7 & 79.2\ensuremath{\pm}0.9 & 197.0\ensuremath{\pm}26.9 & 187.5\ensuremath{\pm}33.0 \\
Zhang et al. & 74.8\ensuremath{\pm}1.6 & 77.4\ensuremath{\pm}2.1 & 325.3\ensuremath{\pm}30.1 & 318.7\ensuremath{\pm}82.3 \\
RayFronts & 60.1\ensuremath{\pm}2.2 & 60.9\ensuremath{\pm}2.4 & 691.9\ensuremath{\pm}142.5 & 707.6\ensuremath{\pm}166.0 \\
ActiveSGM & 82.3\ensuremath{\pm}1.7 & 84.4\ensuremath{\pm}1.9 & 313.7\ensuremath{\pm}76.4 & 301.3\ensuremath{\pm}109.4 \\
GLEAM & 90.0\ensuremath{\pm}0.5 & 91.7\ensuremath{\pm}1.2 & 47.2\ensuremath{\pm}20.0 & 46.4\ensuremath{\pm}11.4 \\
\midrule
\textbf{Ours} & \textbf{94.9\ensuremath{\pm}0.8} & \textbf{97.1\ensuremath{\pm}0.4} & \textbf{14.2\ensuremath{\pm}5.8} & \textbf{8.2\ensuremath{\pm}3.6} \\
\bottomrule
\end{tabular*}
}
\caption{Comparison with state-of-the-art rule- and learning-based methods using two vision-language feature extractors. Ours achieves the highest OCR and consistent performance across the three scale categories.}
\label{tab:main_results}
\end{table*}

\section{Experiments}

\subsection{Implementation Details}

\subsubsection{Dataset.} All experiments are conducted in our simulated urban environments. We use 16 scenes as the training set, 4 as the validation set, and 60 scenes not used during training as the test set. The map size is configured as $200\,\text{m} \times 200\,\text{m}$ for every scene.

\subsubsection{Metrics.} We group classes by volume into $\mathcal{C}=\{c_{\text{small}},c_{\text{medium}},c_{\text{large}}\}$ using ranges $[0.01,5)$, $[5,100)$, and $[100,208.47\mathrm{k}]\,\text{m}^3$, respectively. Let $y_v$ and $\hat{y}_v$ be the ground-truth and reconstructed labels at grid $v$, and $\mathcal{V}_c=\{v\mid y_v\in c\}$. We define
\begin{equation}
    \mathrm{CCR}_c=\frac{\sum_{v\in\mathcal{V}_c}\mathbb{I}[\hat{y}_v=y_v]}{|\mathcal{V}_c|},
    \qquad
    \mathrm{OCR}=\frac{1}{|\mathcal{C}|}\sum_{c\in\mathcal{C}}\mathrm{CCR}_c.
\end{equation}
Unexplored and incorrectly labeled grids contribute zero. We report these ratios as percentages and use \textbf{Var} for the variance among the three CCRs.

\subsubsection{Methods.} We compare state-of-the-art semantic mapping methods. 1) \textbf{Uncertainty \cite{lee2022uncertainty}.} This method selects the next best position by minimizing geometric uncertainty. We equip it with the semantic module to perform semantic mapping. 2) \textbf{Zhang et al. \cite{zhang2024active}.} We apply the semantic uncertainty calculation method from this work to select the next agent pose that minimizes uncertainty. 3) \textbf{RayFronts \cite{alama2025rayfronts}.} This method performs semantic mapping based on frontier-based exploration (FBE). 4) \textbf{ActiveSGM \cite{chen2025understanding}.} We apply the exploration policy from this work by jointly calculating geometric and semantic uncertainty. 5) \textbf{GLEAM \cite{chen2025GLEAM}} is a state-of-the-art RL-based mapping method. We use the semantic reward to match our task setting. All agents share the same pose and camera configuration. We test CLIP \cite{radford2021learning} and DINOv3 with its official dino.txt text-alignment head \cite{simeoni2025dinov3}. Each model uses its paired visual and text encoders, and their normalized cosine similarities form semantic likelihoods. The maximum number of execution steps for each agent is 384. The input global map resolution is $256\times 256$. Three random seeds are used for learning-based agents. For testing, three random initial positions are used for all agents. 

\subsubsection{Training.} Our method is trained end-to-end from scratch with PPO. We use a three-layer ResNet as the feature extractor, a batch size of 256, and a learning rate of $10^{-4}$. The latent dimensions of $Z_\alpha$ and $Z_\beta$ are both 256, and the CLUB predictor is a 256-to-128 MLP. We set $k=0.1$ in Eq.~\ref{eq:class_loss}. The agent is trained for $10^3$ episodes. Training is performed on a single RTX 4090 and takes about 19 hours.

\subsection{Main Results}

Tab.~\ref{tab:main_results} reports the main results. Rule-based methods lag behind learning-based agents, especially on $c_{\text{small}}$, because small objects are reliably mapped only from a narrow range of viewpoints. Learning-based baselines improve exploration through interaction, but still depend on raw VLM likelihoods and remain sensitive to observations from suboptimal ranges. In contrast, our calibration policy mitigates the effects of unreliable likelihoods while the motion policy searches for effective viewpoints, leading to the best OCR and lowest Var under both CLIP and DINOv3 likelihoods.

\subsection{Ablation Studies}

We conduct ablation studies on the proposed modules and report the results in Tab.~\ref{tab:ablation_studies}. Adding LC improves the baseline by enabling adaptive likelihood calibration, especially for small objects. Adding MV without gradient balancing is unstable because the category-wise objectives conflict, while PO restores balanced optimization and substantially reduces Var. Adding MI regularization further improves mapping performance, providing task-level evidence that dependence-regularized representations benefit joint policy learning. Combining all components achieves the best OCR and the lowest Var.

\begin{table}[t]
\centering
{\small
\setlength{\tabcolsep}{0.5pt}
\begin{tabular*}{\columnwidth}{@{\extracolsep{\fill}}c c c c c c c c c@{}}
\toprule
\multicolumn{4}{c}{\textbf{Components}} & \multicolumn{3}{c}{\textbf{CCR (\%)} $\uparrow$} & \multirow{2}{*}{\textbf{OCR (\%)} $\uparrow$} & \multirow{2}{*}{\textbf{Var} $\downarrow$} \\ 
\cmidrule(r){1-4} \cmidrule(r){5-7}
LC & MV & PO & MI & $c_{\text{small}}$ & $c_{\text{medium}}$ & $c_{\text{large}}$ & & \\
\midrule
-- & -- & -- & -- & 82.0 & 97.4 & 99.1 & 92.8 & 59.5  \\
$\checkmark$ & -- & -- & -- & 84.3 & 96.8 & 99.2 & 93.4 & 42.9 \\
$\checkmark$ & $\checkmark$ & -- & -- & 50.2 & 86.5 & 99.4 & 78.7 & 433.9 \\
$\checkmark$ & $\checkmark$ & $\checkmark$ & -- & 92.1 & 97.5 & \textbf{99.8} & 96.5 & 10.4 \\
$\checkmark$ & -- & -- & $\checkmark$ & 88.3 & 96.2 & 99.7 & 94.7 & 22.7 \\
\midrule
$\checkmark$ & $\checkmark$ & $\checkmark$ & $\checkmark$ & \textbf{93.5} & \textbf{99.6} & 99.7 & \textbf{97.6} & \textbf{8.4} \\
\bottomrule
\end{tabular*}
}
\caption{Ablation studies on the main components. LC denotes likelihood calibration, MV denotes multi-value prediction, PO denotes Pareto optimization, and MI denotes mutual-information dependence regularization. The row with all components enabled is the full model.}
\label{tab:ablation_studies}
\end{table}

\subsection{Hyperparameter Study}

Tab.~\ref{tab:hyperparameter} studies the sensitivity to the MI regularization weight $k$. When $k$ is small, the penalty is weak and performance remains close to the Pareto-only setting in Tab.~\ref{tab:ablation_studies}. With a moderate $k$, mapping performance improves and multi-scale variance decreases, showing the benefit of balancing task optimization and the estimated-MI penalty. When $k$ is too large, this penalty dominates the update and degrades performance. We select $k$ by the highest validation-set reward and fix it for testing.

\begin{table}[t]
\centering
{\small
\setlength{\tabcolsep}{1.2pt}
\begin{tabular*}{\columnwidth}{@{\extracolsep{\fill}}c c c c c c@{}}
\toprule
\multirow{2}{*}{$k$} & \multicolumn{3}{c}{\textbf{CCR (\%)} $\uparrow$} & \multirow{2}{*}{\textbf{OCR (\%)} $\uparrow$} & \multirow{2}{*}{\textbf{Var} $\downarrow$} \\
\cmidrule(r){2-4}
& $c_{\text{small}}$ & $c_{\text{medium}}$ & $c_{\text{large}}$ & & \\
\midrule
0.01 & 91.0 & 98.5 & 99.8 & 96.4 & 14.9 \\
0.05 & 92.7 & 98.5 & 99.8 & 97.0 & 9.6 \\
0.10 & \textbf{93.5} & \textbf{99.6} & 99.7 & \textbf{97.6} & \textbf{8.4} \\
0.15 & 90.4 & 98.5 & 99.8 & 96.2 & 17.3 \\
0.20 & 87.8 & 96.1 & \textbf{99.9} & 94.6 & 25.4 \\
\bottomrule
\end{tabular*}
}
\caption{Sensitivity to the MI regularization weight $k$.}
\label{tab:hyperparameter}
\end{table}

\subsection{Performance Analysis}

\begin{figure*}[t]
    \centering
    \includegraphics[width=0.75\textwidth]{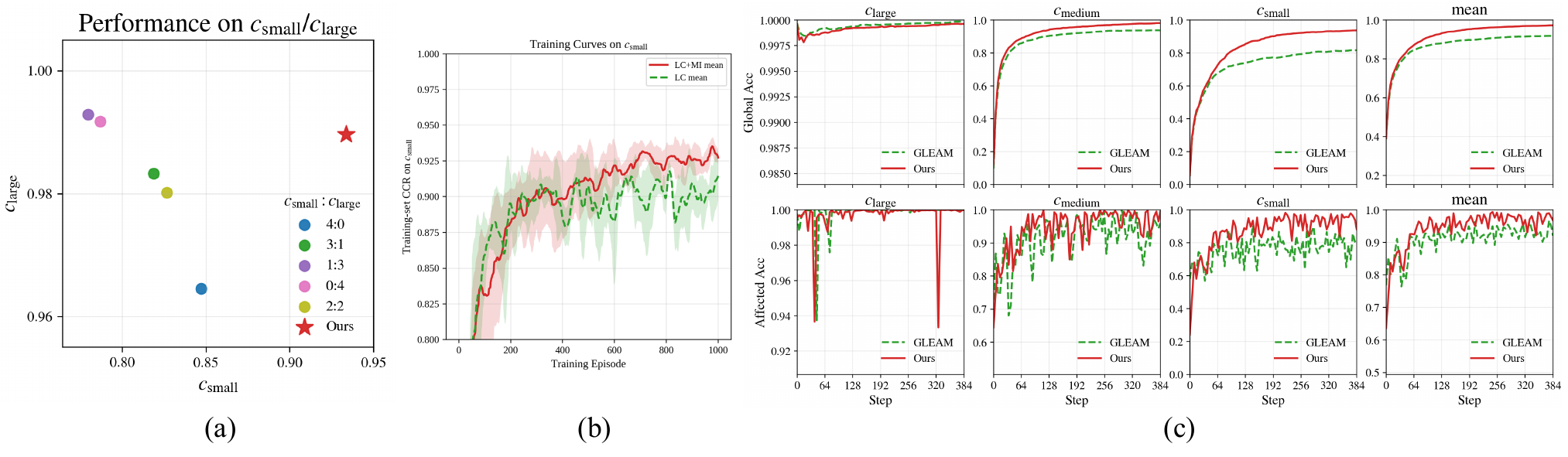}
    \caption{(a) Performance on $c_{\text{small}}$ and $c_{\text{large}}$ under different reward allocations; ours is more balanced and closer to the Pareto frontier. (b) Training-set CCR on $c_{\text{small}}$; MI stabilizes training and improves final performance. (c) Step-level test performance. The rows show global-map and affected-grid accuracy; ours achieves faster coverage and more accurate updates than GLEAM.}
    \label{fig:pareto_acc}
\end{figure*}

\begin{figure*}[t]
    \centering
    \includegraphics[width=0.6\textwidth]{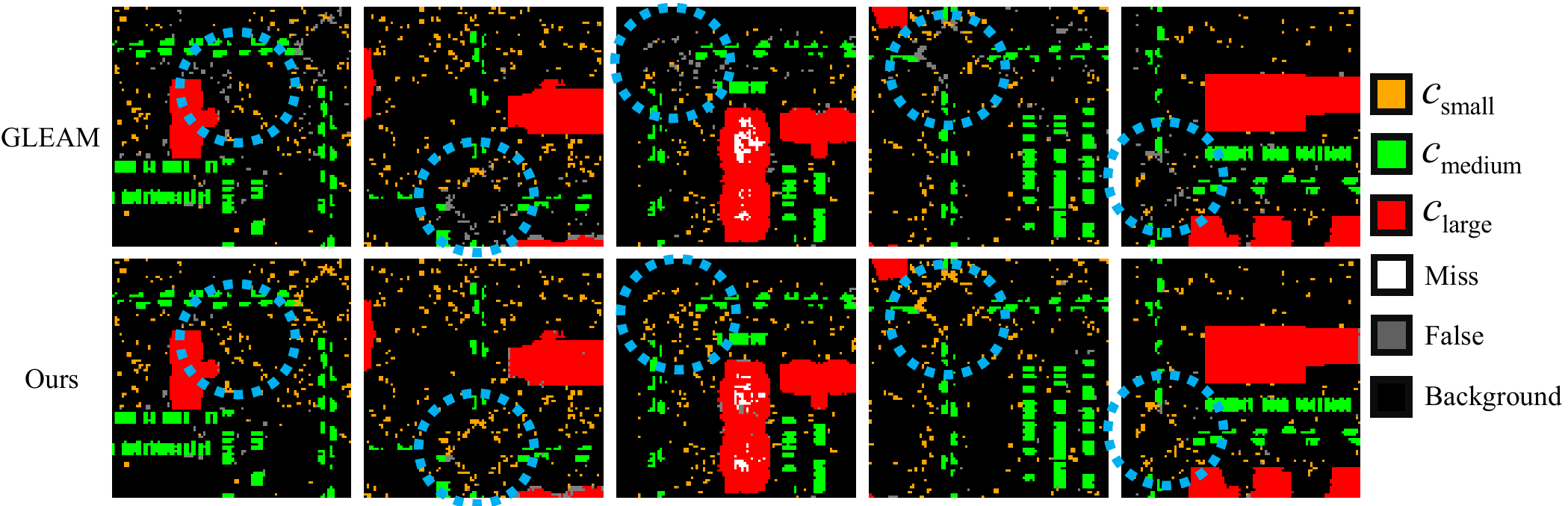}
    \caption{Qualitative semantic maps from ours and GLEAM. Colors denote correct classes; white, gray, and black denote unexplored regions, errors, and background. GLEAM misses small objects frequently, whereas ours maintains consistent multi-scale coverage.}
    \label{fig:reconstruct}
\end{figure*}

\subsubsection{Pareto Frontier.}
We study the Pareto frontier by varying the reward allocation between $c_{\text{small}}$ and $c_{\text{large}}$, using LC as the baseline. As shown in Fig.~\ref{fig:pareto_acc}(a), our method achieves the most balanced performance across the two classes, indicating that Pareto optimization alleviates gradient conflicts and moves the policy toward the frontier.

\subsubsection{Dependence Regularization.}
Fig.~\ref{fig:pareto_acc}(b) compares the training curves of LC and LC with MI on $c_{\text{small}}$. MI stabilizes training after about 500 episodes and reaches higher final performance, while LC fluctuates around a lower mean performance. This shows that MI regularization supports more stable and effective learning.

\subsubsection{Likelihood Calibration.}
We compare fixed likelihood scaling with our learned perception policy trained using LC and MI. MI serves only as a training-time regularizer, and its CLUB predictor is discarded after training; it therefore introduces no additional inference-time modules, parameters, or computation. We disable Pareto optimization in this comparison. As shown in Tab.~\ref{tab:manual_comparison}, context-adaptive likelihood calibration with MI dependence regularization consistently outperforms fixed global weights, showing the effectiveness of the learned calibration.

\begin{table}[H]
\centering
{\small
\setlength{\tabcolsep}{0.5pt}
\begin{tabular*}{\columnwidth}{@{\extracolsep{\fill}}c c c c c c@{}}
\toprule
\multirow{2}{*}{\textbf{Method}} & \multicolumn{3}{c}{\textbf{CCR (\%)} $\uparrow$} & \multirow{2}{*}{\textbf{OCR (\%)} $\uparrow$} & \multirow{2}{*}{\textbf{Var} $\downarrow$} \\ 
\cmidrule(r){2-4} 
& $c_{\text{small}}$ & $c_{\text{medium}}$ & $c_{\text{large}}$ & & \\
\midrule
GLEAM-0.2 & 78.9 & 92.4 & 98.4 & 89.9 & 66.2 \\
GLEAM-0.6 & 80.6 & 93.2 & 99.2 & 91.0 & 60.2 \\
GLEAM-1.0 & 83.7 & 93.3 & 98.8 & 91.9 & 38.8 \\
GLEAM-1.4 & 80.3 & 94.5 & 99.4 & 91.4 & 65.5 \\
GLEAM-1.8 & 76.3 & 95.8 & 98.5 & 90.2 & 97.4 \\
\midrule
\textbf{Ours} & \textbf{88.3} & \textbf{96.2} & \textbf{99.7} & \textbf{94.7} & \textbf{22.7} \\
\bottomrule
\end{tabular*}
}
\caption{Fixed likelihood scaling versus learned adaptive calibration trained with LC and MI. `GLEAM-$x$' uses a fixed likelihood weight of $x$. MI introduces no additional inference-time modules or parameters.}
\label{tab:manual_comparison}
\end{table}

\subsubsection{Step-level Performance.}
We further compare step-level performance with GLEAM by measuring global-map accuracy and affected-grid accuracy after each update. Fig.~\ref{fig:pareto_acc}(c) shows that our method improves mapping speed and local update accuracy, validating the effectiveness of calibration.

\subsection{Visualization}

Fig.~\ref{fig:reconstruct} visualizes reconstructed semantic maps. Compared with GLEAM, our method reduces missed small objects and incomplete exploration, producing more consistent maps across object scales.

\section{Conclusion}

In this paper, we propose Multi-Scale Semantic Mapping, which differs from existing semantic mapping tasks by introducing objects with significant size divergence. 
We build a simulated urban dataset using real-world city layouts with LLM planning and high-fidelity rendering to model real-world urban conditions and large object-scale variations. 
To mitigate erroneous likelihoods caused by suboptimal observations in large simulated urban environments, we introduce a likelihood calibration module that predicts map updating strengths, thereby improving mapping accuracy. 
To mitigate performance drop caused by potential calibration and motion policy dependence, we penalize their estimated representation mutual information during training, encouraging complementary behavior learning.
To learn the heterogeneous mapping strategies required by objects
at different scales, we use category-wise value heads to model the complex policy advantages, and use Pareto optimization to balance the gradient conflicts. 
Experimental results show that our method outperforms existing rule-based and learning-based methods, especially on small-scale objects. Since none of existing methods consider dynamic semantic mapping currently, we do not aim to solve this challenge setting in this work. Future work will extend the method to dynamic objects.

\bibliography{sample}
\bibliographystyle{plain}

\clearpage


\appendix

\setlength{\abovedisplayskip}{6pt plus 2pt minus 2pt}
\setlength{\belowdisplayskip}{6pt plus 2pt minus 2pt}
\setlength{\abovedisplayshortskip}{3pt plus 2pt}
\setlength{\belowdisplayshortskip}{4pt plus 2pt minus 1pt}
\setlength{\textfloatsep}{8pt plus 2pt minus 2pt}
\setlength{\floatsep}{8pt plus 2pt minus 2pt}
\setlength{\intextsep}{6pt plus 2pt minus 2pt}
\setlength{\dbltextfloatsep}{5pt plus 1pt minus 1pt}
\setlength{\dblfloatsep}{4pt plus 1pt minus 1pt}
\captionsetup{skip=4pt}
\setcounter{dbltopnumber}{3}
\renewcommand{\dbltopfraction}{0.95}
\renewcommand{\textfraction}{0.03}
\makeatletter
\setlength{\@fptop}{0pt}
\setlength{\@dblfptop}{0pt}
\makeatother

\begin{bibunit}[plain]

\section{Dataset}

\subsection{GIS Data Acquisition}

We first select specific locations from the OSM system and cut maps in fixed size. For each selected area of interest, we extract the street network and building areas. Since the raw OSM data may not be connected, we mannually fix the raw data to ensure connectivity. The fixed data is orgnized into a graph: each node represents a block or a road segementation, and eages represent the connections of adjacent nodes. This infomation represents the spatial relationships between different urban areas and can be processed by LLMs.  The OSM data contain road and building names, and we gather these information along with the graph structure for the next LLM planning stage.

\subsection{LLM Asset Design}
The first-stage LLM receives the scene and block types summarized from the GIS collection together with the repository asset catalog. It produces a JSON specification of a reusable CGA library, including the applicable scene and block types, object assets, their spatial-distribution functions, and the parameters exposed by these functions. The prompt is reproduced in Tab.~\ref{tab:asset_design_prompt}.

\begin{table}[H]
\small
\setlength{\tabcolsep}{3pt}
\begin{tabularx}{\columnwidth}{@{}X@{}}
\toprule
{\itshape
You are an intelligent agent to plan a city layput using CityEngine.

\medskip
You will receive a list of scene types and block types, and also a list of available object asset names.

\medskip
You need to design a CGA library for CityEngine. Each CGA should contain these informations:

\medskip
1. Scene type.\newline
2. Block type.\newline
3. Available object asset names.\newline
4. Distribution functions for each object.\newline
5. Parameters for each distribution function.

\medskip
Do not use scenes, blocks and assets out of the provided lists. Think step by step of the distribution design and output your thought. Output a json file that contains all the information of the CGA library.

\medskip
Scene type: \texttt{\{scene\_types\}}\newline
Block type: \texttt{\{block\_types\}}\newline
Object assets: \texttt{\{object\_assets\}}
} \\
\bottomrule
\end{tabularx}
\caption{Prompt for LLM asset design.}
\label{tab:asset_design_prompt}
\end{table}

\subsection{LLM Scene Planning}
For each scene, the second-stage LLM receives a JSON description containing its scene type and block information together with the available CGA rules. It assigns a CGA rule and the parameters of its distribution functions to every block while considering the block type and its surroundings. The prompt is reproduced in Tab.~\ref{tab:scene_planning_prompt}.

\begin{table}[H]
\small
\setlength{\tabcolsep}{3pt}
\begin{tabularx}{\columnwidth}{@{}X@{}}
\toprule
{\itshape
You are an intelligent agent to plan a city layput using CityEngine.

\medskip
You will receive a json file of current scene that contains its type with block information, and a list of CGA rules.

\medskip
You need to assign the CGA rules to each block, and assign the parameters of the distribution functions in the rules. Think step by step the type, consider its block type, and the surrounding blocks. Think about how real world objects ditribute and make sure that the parameters are aligned with real world.

\medskip
Scene info path: \texttt{\{scene\_info\_path\}}\newline
CGA library path: \texttt{\{cga\_library\_path\}}
} \\
\bottomrule
\end{tabularx}
\caption{Prompt for LLM scene planning.}
\label{tab:scene_planning_prompt}
\end{table}

\subsection{Scene generation.}
With the planned scene road graph and CGA assignments, CityEngine uses these information and generate the 3D scene. It is then exported to Isaac Sim for training or testing.

\section{Method}

\subsection{Likelihood Generation}

We describe the raw VLM semantic likelihood generation process in this section.

\textbf{Raw likelihood generation.} Given an RGB image $I_t$, we extract patch-level visual features and class text features with a VLM. For class $c$, multiple prompts are allowed (indexed by $m$). Let $f_v(\cdot)$ and $f_t(\cdot)$ denote visual and text encoders, and let $A_v(\cdot)$ and $A_t(\cdot)$ denote their alignment projections into the same feature space. These projections are identities for an already aligned VLM. For patch $i$, the normalized visual and text embeddings are:
\begin{equation}
\begin{aligned}
\tilde{\mathbf{v}}_{t,i} &= A_v\!\left(f_v(I_t)_i\right), 
\hat{\mathbf{v}}_{t,i} = \frac{\tilde{\mathbf{v}}_{t,i}}{\lVert \tilde{\mathbf{v}}_{t,i} \rVert_2}, \\
\tilde{\mathbf{t}}_{c,m} &= A_t\!\left(f_t(p_{c,m})\right),
\hat{\mathbf{t}}_{c,m} = \frac{\tilde{\mathbf{t}}_{c,m}}{\lVert \tilde{\mathbf{t}}_{c,m} \rVert_2}.
\end{aligned}
\end{equation}
Prompt-level cosine similarity is computed as:
\begin{equation}
s_{t,i,c,m} = \hat{\mathbf{v}}_{t,i}^\top \hat{\mathbf{t}}_{c,m}.
\end{equation}
If class $c$ has multiple prompts, we aggregate them by max pooling:
\begin{equation}
s_{t,i,c} = \max_{m \in \{1,\dots,M_c\}} s_{t,i,c,m}.
\end{equation}
The patch-level class similarity map is then resized to image resolution:
\begin{equation}
S_{t,c}(u,v) = \mathrm{Interp}\!\left(\{s_{t,i,c}\}; H, W\right).
\end{equation}
The raw pixel-level class likelihood is:
\begin{equation}
P_t^{\mathrm{raw}}(y=c \mid u,v) =
\frac{\exp\!\left(S_{t,c}(u,v)\right)}
{\sum_{k=1}^{C}\exp\!\left(S_{t,k}(u,v)\right)}.
\end{equation}

\textbf{Depth Back-Projection to 3D.} Let $d_{uv}$ be depth at pixel $(u,v)$, and let camera intrinsics be
\begin{equation}
\mathbf{K}=\begin{bmatrix}
f_x & 0 & c_x\\
0 & f_y & c_y\\
0 & 0 & 1
\end{bmatrix},
\qquad
\mathbf{p}_{uv}=\begin{bmatrix}
u\\v\\1\end{bmatrix}.
\end{equation}
The 3D point in camera coordinates is:
\begin{equation}
\mathbf{x}^{c}_{uv}=d_{uv}\mathbf{K}^{-1}\mathbf{p}_{uv}.
\end{equation}
Using homogeneous coordinates, world coordinates are:
\begin{equation}
\tilde{\mathbf{x}}^{w}_{uv}
=
\mathbf{T}_{cw}
\begin{bmatrix}
\mathbf{x}^{c}_{uv}\\1
\end{bmatrix},
\end{equation}
where $\mathbf{T}_{cw}$ is the camera-to-world transform.

\textbf{Voxel Aggregation and Global Likelihood.}
Let $\mathcal{P}_{t,x,y,z}$ be pixels whose points are projected to the same 3D cell $(x,y,z)$. We compute the voxel-level raw VLM logit of the current observation by averaging the aligned similarities:
\begin{equation}
l_{t,x,y,z,c} = \frac{1}{|\mathcal{P}_{t,x,y,z}|} \sum_{(u_p,v_p)\in\mathcal{P}_{t,x,y,z}} S_{t,c}(u_p,v_p).
\end{equation}
Before calibration, the current logits are projected to a raw 2D local map. Let $\mathcal{Z}^{\mathrm{obs}}_{t,x,y}$ be the height bins observed at planar cell $(x,y)$ and let $\mathrm{Softmax}_c$ act along the class dimension. The channel-first observation map is
\begin{equation}
\begin{aligned}
r_{t,x,y,c}^{\mathrm{obs}}
&=\frac{1}{|\mathcal{Z}^{\mathrm{obs}}_{t,x,y}|}
\sum_{z\in\mathcal{Z}^{\mathrm{obs}}_{t,x,y}}l_{t,x,y,z,c},\\
\mathbf m_t^{\mathrm{obs}}
&=\mathrm{Permute}\!\left(\mathrm{Softmax}_c[\mathbf r_t^{\mathrm{obs}}]\right).
\end{aligned}
\end{equation}
with unobserved cells zero-filled. As detailed below, both policies receive the pre-update context $z_t^{-}$ containing the previous global map $\mathbf M_{t-1}$ and current observation $\mathbf m_t^{\mathrm{obs}}$. The calibration policy first predicts $\boldsymbol{\beta}_t\sim\pi_\theta^\beta(\cdot\mid z_t^{-})$. Writing $v=(x,y,z)$, the calibrated likelihood used in Eq.~\ref{eq:parameterization} is
\begin{equation}
P_t^{\mathrm{cal}}(c\mid z_t^{-},v)
=
\frac{\exp\!\left(\beta_{t,v,c}l_{t,v,c}\right)}
{\sum_{j=1}^{C}\exp\!\left(\beta_{t,v,j}l_{t,v,j}\right)}.
\end{equation}
The calibrated likelihood is then fused with the previous global log-odds state, as in Eq.~\ref{eq:updating}:
\begin{equation}
L_{v,c}^{t}
=L_{v,c}^{t-1}
+\log\!\left(\frac{P_t^{\mathrm{cal}}(c\mid z_t^{-},v)}{1-P_t^{\mathrm{cal}}(c\mid z_t^{-},v)}\right).
\end{equation}
Finally, let $\mathcal{Z}_{t,x,y}$ be the valid height bins at planar cell $(x,y)$. We project the updated voxel log-odds and form the channel-first global map:
\begin{equation}
\begin{aligned}
g_{t,x,y,c}
&=\frac{1}{|\mathcal{Z}_{t,x,y}|}
\sum_{z\in\mathcal{Z}_{t,x,y}}L_{(x,y,z),c}^{t},\\
P_t^{\mathrm{glob}}(x,y,c)
&=\frac{\exp(g_{t,x,y,c})}{\sum_{j=1}^{C}\exp(g_{t,x,y,j})},\\
\mathbf M_t
&=\mathrm{Permute}(\mathbf P_t^{\mathrm{glob}})
\in\mathbb R^{C\times H\times W}.
\end{aligned}
\end{equation}
Thus, the causal order is $(\mathbf M_{t-1},\mathbf m_t^{\mathrm{obs}})\rightarrow\boldsymbol\beta_t\rightarrow\mathbf M_t$, or equivalently $\mathbf M_t=\mathcal F(\mathbf M_{t-1},\mathbf m_t^{\mathrm{obs}};\boldsymbol\beta_t)$. In particular, $\mathbf M_t$ is not used to predict $\boldsymbol\beta_t$; it becomes the previous global map at step $t+1$.

\subsection{Network Input}
We adopt a two-branch encoder containing a historical-pose branch and a semantic-map branch. The historical-pose branch is:
\begin{equation}
\begin{aligned}
\mathbf{p}_t &= [x_t, y_t, z_t, \phi_t, \theta_t, \psi_t], \\
\mathbf{S}_t &= [\mathbf{p}_{t-L+1}, \ldots, \mathbf{p}_t] \in \mathbb{R}^{L \times 6}.
\end{aligned}
\end{equation}
where $L$ denotes the history length. Before updating $\mathbf M_t$, the map branch concatenates the previous global map with the current raw observation map. The map input and pre-update policy context are
\begin{equation}
\begin{aligned}
\mathbf X_t^{\mathrm{map}}
&=\mathrm{Concat}\!\left(\mathbf M_{t-1},\mathbf m_t^{\mathrm{obs}}\right)
\in\mathbb R^{2C\times H\times W},\\
z_t^{-}
&=\left\{\mathbf S_t,\mathbf X_t^{\mathrm{map}}\right\}
=\left\{\mathbf S_t,\mathbf M_{t-1},\mathbf m_t^{\mathrm{obs}}\right\}.
\end{aligned}
\end{equation}
The previous observations are recursively summarized by $\mathbf M_{t-1}$, while $\mathbf m_t^{\mathrm{obs}}$ preserves the current uncalibrated evidence. The shared encoder processes $z_t^{-}$, and the motion and calibration branches jointly predict $a_t$ and $\boldsymbol\beta_t$. The latter is then used to update $\mathbf M_{t-1}$ into $\mathbf M_t$ as defined above. This ordering prevents the updated map from being used circularly to predict its own calibration.

\subsection{ELBO Clarification}
In the main paper, $q_{\phi}(Z_\alpha,Z_\beta\mid z_{1:t})$ indicates that the motion and calibration representations are induced by the observation history. In the appendix notation, this history is summarized by the pre-update context $z_t^-$. The conditioning in the encoder therefore specifies representation generation; it does not mean that the implemented regularizer optimizes MI separately for each fixed $z_t^-$. Under the on-policy visitation distribution $d^{\pi_\theta}$, the conditional encoders induce the aggregate distribution
\begin{equation}
\begin{aligned}
\bar q_\phi(Z_\alpha,Z_\beta)
&=\mathbb E_{z_t^-\sim d^{\pi_\theta}}
\left[q_\phi(Z_\alpha,Z_\beta\mid z_t^-)\right],\\
I_{\bar q}(Z_\alpha;Z_\beta)
&=\mathcal D_{\mathrm{KL}}\!\left(
\bar q_{\alpha\beta}\parallel\bar q_\alpha\bar q_\beta
\right).
\end{aligned}
\end{equation}
The implemented CLUB loss estimates this unconditional aggregate MI: same-transition features sample $\bar q_{\alpha\beta}$, whereas cross-sample pairs approximate $\bar q_\alpha\bar q_\beta$.

The PPO objective is related to the ELBO reconstruction term through the category-wise coverage reward. Using $R_{c,t}=\mathrm{CR}_{t+1}^{(c)}-\mathrm{CR}_{t}^{(c)}$, its discounted episode return satisfies
\begin{equation}
\begin{aligned}
\sum_{t=0}^{T-1}\gamma^tR_{c,t}
&=-\mathrm{CR}_{0}^{(c)}
+(1-\gamma)\sum_{t=1}^{T-1}\gamma^{t-1}\mathrm{CR}_{t}^{(c)}\\
&\quad+\gamma^{T-1}\mathrm{CR}_{T}^{(c)}.
\end{aligned}
\end{equation}
For $\gamma=1$, this reduces exactly to $\mathrm{CR}_{T}^{(c)}-\mathrm{CR}_{0}^{(c)}$; for $\gamma<1$, it additionally rewards reaching accurate coverage earlier. Since unexplored and incorrectly labeled grids contribute zero to $\mathrm{CR}_{t}^{(c)}$, PPO optimizes a task-level surrogate for the ELBO reconstruction term $\mathbb E[\log p(M\mid Z_\alpha,Z_\beta)]$, while CLUB regularizes the unconditional dependence of the aggregate representations. Thus, the implemented objective is related to the two ELBO terms.

\subsection{Dependence Regularization}
We use MI as a weighted dependence regularizer rather than imposing statistical independence. We denote the motion feature as $Z_{\alpha}$ and the calibration feature as $Z_{\beta}$. Their realizations at time $t$ are $\mathbf{z}_{\alpha,t}$ and $\mathbf{z}_{\beta,t}$. The shared network input is the pre-update context $z_t^{-}$, and the two branch features are computed as:
\begin{equation}
\begin{aligned}
h_t &= \mathrm{Shared}_\Theta(z_t^{-}), \\
\mathbf{z}_{\alpha,t} &= \mathrm{Norm}\!\left(\mathrm{Proj}_{\alpha}\!\left(\mathrm{Adapter}_{\alpha}(h_t)\right)\right),\\
\mathbf{z}_{\beta,t} &= \mathrm{Norm}\!\left(\mathrm{Proj}_{\beta}\!\left(\mathrm{Adapter}_{\beta}(h_t)\right)\right).
\end{aligned}
\end{equation}
where $h_t$ is the output of the shared feature extractor.

The CLUB estimator models a conditional Gaussian distribution:
\begin{equation}
q_{\alpha \to \beta}(Z_{\beta}\mid Z_{\alpha})
= \mathcal{N}\!\left(\mu_{\alpha\to\beta}(Z_{\alpha}),\,\mathrm{diag}(\sigma^2_{\alpha\to\beta}(Z_{\alpha}))\right),
\end{equation}
where $\log \sigma^2$ is clamped for numerical stability. The positive-pair log-likelihood is
\begin{equation}
\log q_{\alpha\to\beta}(\mathbf{z}_{\beta,t}\mid \mathbf{z}_{\alpha,t}),
\end{equation}
and the negative-pair term is approximated using $K$ features drawn from other samples in the batch:
\begin{equation}
\frac{1}{K}\sum_{k=1}^{K} \log q_{\alpha\to\beta}(\mathbf{z}_{\beta,t,k}^{-}\mid \mathbf{z}_{\alpha,t}).
\end{equation}
Thus, the variational CLUB estimate is:
\begin{equation}
\mathcal U_{\alpha\to\beta}
=
\mathbb E\!\left[\log q_{\alpha\to\beta}(Z_{\beta}\mid Z_{\alpha})\right]
-
\mathbb E\!\left[\log q_{\alpha\to\beta}(Z_{\beta}^{-}\mid Z_{\alpha})\right].
\end{equation}
The CLUB-based dependence penalty used during the policy-feature update is:
\begin{equation}
\mathcal L_{\mathrm{MI}}
=
\mathcal U_{\alpha\to\beta}.
\end{equation} 
We use an alternating training procedure. Before each policy-feature update, the estimator is first optimized for three steps using negative log-likelihood:
\begin{equation}
\mathcal L_{\mathrm{estimator}}
=
-\mathbb E[\log q_{\alpha\to\beta}(Z_{\beta}\mid Z_{\alpha})].
\end{equation}
The estimator parameters are then frozen, and the policy feature extractor is optimized by the policy objective augmented with the weighted $\mathcal L_{\mathrm{MI}}$ term. This penalty discourages excessive predictability between the two policy features while retaining task-relevant shared information; it neither enforces independence nor eliminates all coupling.

\subsection{Category-Wise PPO Objective}
We present the class-wise PPO loss in this section. Let $c \in \{1,\dots,C\}$ denote the category index, and define the transition reward consistently with Eq.~\ref{eq:reward} as $R_{c,t}=\mathrm{CR}_{t+1}^{(c)}-\mathrm{CR}_{t}^{(c)}$. The class-wise generalized advantage estimate is computed from the TD residuals
\begin{equation}
\delta_{c,t}
=R_{c,t}+\gamma(1-d_t)V_{c,\psi_{\mathrm{old}}}(h_{t+1})
-V_{c,\psi_{\mathrm{old}}}(h_t),
\end{equation}
where $d_t$ indicates whether the transition terminates the episode. For a rollout ending at step $T$, the multi-step advantage is
\begin{equation}
\hat{A}_{t,c} = \sum_{l=0}^{T-t-1} (\gamma \lambda)^l \delta_{c,t+l}.
\end{equation}
Here, $V_{c,\psi}(h_t)$ is the value head for category $c$, and $\gamma$ and $\lambda$ are the discount factor and GAE trace coefficient. The corresponding GAE return used as the value regression target is
\begin{equation}
V_{c,t}^{\mathrm{targ}}
=\hat A_{t,c}+V_{c,\psi_{\mathrm{old}}}(h_t).
\end{equation}
Let the joint policy conditioned on the pre-update context factorize into the motion and likelihood-calibration policies:
\begin{equation}
\pi_\theta(a_t,\boldsymbol{\beta}_t\mid z_t^{-})
=\pi_\theta^\alpha(a_t\mid Z_{\alpha,t})
\pi_\theta^\beta(\boldsymbol{\beta}_t\mid Z_{\beta,t}),
\end{equation}
where $\boldsymbol{\beta}_t$ collects the spatial-semantic calibration outputs. The importance ratio is
\begin{equation}
r_t(\theta)
=
\frac{
\pi_\theta^\alpha(a_t\mid Z_{\alpha,t})
\pi_\theta^\beta(\boldsymbol{\beta}_t\mid Z_{\beta,t})}
{
\pi_{\theta_{\mathrm{old}}}^\alpha(a_t\mid Z_{\alpha,t})
\pi_{\theta_{\mathrm{old}}}^\beta(\boldsymbol{\beta}_t\mid Z_{\beta,t})}.
\end{equation}
The clipped value prediction is
\begin{equation}
\begin{aligned}
\Delta V_{c,t} &= V_{c,\psi}(h_t)-V_{c,\psi_{\mathrm{old}}}(h_t),\\
V_{c,t}^{\mathrm{clip}}
&=V_{c,\psi_{\mathrm{old}}}(h_t)
+\operatorname{clip}(\Delta V_{c,t},-\epsilon_v,\epsilon_v).
\end{aligned}
\end{equation}
For compactness, define the clipped ratio, policy surrogate, and value residuals as
\begin{align}
\bar r_t(\theta)
&=\operatorname{clip}(r_t(\theta),1-\epsilon,1+\epsilon),\\
s_{t,c}(\theta)
&=\min\!\left(r_t(\theta)\hat A_{t,c},
\bar r_t(\theta)\hat A_{t,c}\right),\\
e_{t,c}&=V_{c,\psi}(h_t)-V_{c,t}^{\mathrm{targ}},\\
\bar e_{t,c}&=V_{c,t}^{\mathrm{clip}}-V_{c,t}^{\mathrm{targ}}.
\end{align}
The policy, value, and entropy losses are
\begin{align}
\mathcal L_{\mathrm{policy}}^{(c)}
&=-\hat{\mathbb E}_t[s_{t,c}(\theta)],\\
\mathcal L_{\mathrm{value}}^{(c)}
&=\hat{\mathbb E}_t[\max(e_{t,c}^2,\bar e_{t,c}^2)],\\
\mathcal L_{\mathrm{entropy}}
&=-\hat{\mathbb E}_t\!\left[
\mathcal H_t^\alpha+\mathcal H_t^\beta
\right].
\end{align}
Here, $\mathcal H_t^j=\mathcal H(\pi_\theta^j(\cdot\mid Z_{j,t}))$ for $j\in\{\alpha,\beta\}$.
The category-specific PPO loss is then
\begin{equation}
\label{eq:ppo_final}
\mathcal L_{\mathrm{PPO}}^{(c)}
=\mathcal L_{\mathrm{policy}}^{(c)}
+k_1\mathcal L_{\mathrm{value}}^{(c)}
+\frac{k_2}{C}\mathcal L_{\mathrm{entropy}},
\end{equation}
where $\psi$ denotes the value-head parameters, $\epsilon_v$ is the value-clipping threshold, and $k_1$ and $k_2$ weight the value and entropy terms.

\section{Experiments}

\subsection{Additional Details}

Unless noted, DINOv3 is the VLM. Each calibration coefficient $\beta_{t,v,c}$ is a discrete action with support $\mathcal B=\{0.2,0.4,\ldots,1.8\}$. For every spatial-semantic entry, $\pi_\theta^\beta$ predicts a categorical distribution over these nine values, and the PPO importance ratio uses the categorical log-probability of the selected value. For a controlled comparison, the manual calibration factors of GLEAM in Tab.~\ref{tab:manual_comparison} use the same support, from $0.2$ to $1.8$. We set $(\gamma,\lambda,\epsilon,\epsilon_v,k_1,k_2)=(0.99,0.95,0.2,0.2,0.8,0.005)$. These settings are fixed across variants without separate tuning.

\subsection{Additional Evaluation Metrics}

We report mean area under the ROC curve (mAUC), mean intersection over union (mIoU), and F-1 score.

\begin{table*}[t]
\centering
{\small
\setlength{\tabcolsep}{2.0pt}
\begin{tabular*}{\textwidth}{@{\extracolsep{\fill}}l c c c c c c@{}}
\toprule
\multirow{2}{*}{\textbf{Method}} & \multicolumn{2}{c}{\textbf{mAUC (\%)} $\uparrow$} & \multicolumn{2}{c}{\textbf{mIoU (\%)} $\uparrow$} & \multicolumn{2}{c}{\textbf{F-1 (\%)} $\uparrow$} \\
\cmidrule(r){2-3} \cmidrule(r){4-5} \cmidrule(r){6-7}
& CLIP & DINOv3 & CLIP & DINOv3 & CLIP & DINOv3 \\
\midrule
Uncertainty & 94.3\ensuremath{\pm}1.1 & 96.8\ensuremath{\pm}0.2 & 70.2\ensuremath{\pm}1.8 & 72.6\ensuremath{\pm}3.2 & 80.1\ensuremath{\pm}4.3 & 82.2\ensuremath{\pm}2.8 \\
Zhang et al. & 94.9\ensuremath{\pm}2.0 & 96.4\ensuremath{\pm}0.9 & 68.0\ensuremath{\pm}1.4 & 70.1\ensuremath{\pm}0.5 & 78.9\ensuremath{\pm}1.8 & 79.8\ensuremath{\pm}0.9 \\
RayFronts & 87.8\ensuremath{\pm}2.4 & 88.9\ensuremath{\pm}2.7 & 55.3\ensuremath{\pm}3.4 & 59.1\ensuremath{\pm}3.2 & 67.3\ensuremath{\pm}2.8 & 69.0\ensuremath{\pm}3.5 \\
ActiveSGM & 94.5\ensuremath{\pm}0.6 & 96.8\ensuremath{\pm}0.2 & 70.6\ensuremath{\pm}2.0 & 72.2\ensuremath{\pm}2.0 & 79.5\ensuremath{\pm}3.1 & 81.4\ensuremath{\pm}1.7 \\
GLEAM & 95.8\ensuremath{\pm}1.1 & 98.6\ensuremath{\pm}0.1 & 88.6\ensuremath{\pm}2.1 & 90.9\ensuremath{\pm}1.0 & 93.7\ensuremath{\pm}0.7 & 95.1\ensuremath{\pm}0.5 \\
\midrule
\textbf{Ours} & \textbf{98.1\ensuremath{\pm}2.1} & \textbf{99.4\ensuremath{\pm}0.1} & \textbf{94.4\ensuremath{\pm}1.1} & \textbf{96.1\ensuremath{\pm}0.2} & \textbf{95.9\ensuremath{\pm}1.1} & \textbf{98.0\ensuremath{\pm}0.1} \\
\bottomrule
\end{tabular*}
}
\caption{Additional metrics corresponding to Tab.~\ref{tab:main_results}.}
\label{tab:additional_metrics}
\end{table*}

Let $\mathcal{V}$ denote the evaluated grids, $\mathcal{V}_c$ the grids of category $c$, $\mathcal{V}_{\neg c}=\mathcal{V}\setminus\mathcal{V}_c$, and $p_{v,c}$ the final predicted likelihood. The predicted set is $\widehat{\mathcal{V}}_c=\{v\in\mathcal{V}\mid\hat y_v=c\}$. We compute one-vs-rest ROC-AUC as
{\small
\begin{equation}
\begin{aligned}
\mathrm{AUC}_c
&=\frac{1}{|\mathcal{V}_c|\,|\mathcal{V}_{\neg c}|}
\sum_{v^+\in\mathcal{V}_c}
\sum_{v^-\in\mathcal{V}_{\neg c}}\\[-2pt]
&\quad\left(\mathbb{I}[p_{v^+,c}>p_{v^-,c}]
+\tfrac{1}{2}\mathbb{I}[p_{v^+,c}=p_{v^-,c}]\right),\\[-2pt]
\mathrm{mAUC}
&=\frac{1}{|\mathcal{C}|}\sum_{c\in\mathcal{C}}\mathrm{AUC}_c.
\end{aligned}
\end{equation}
}
For the segmentation metrics, we define
{\small
\begin{equation}
\begin{gathered}
\mathrm{TP}_c=|\widehat{\mathcal{V}}_c\cap\mathcal{V}_c|,\quad
\mathrm{FP}_c=|\widehat{\mathcal{V}}_c\setminus\mathcal{V}_c|,\quad
\mathrm{FN}_c=|\mathcal{V}_c\setminus\widehat{\mathcal{V}}_c|,\\
\mathrm{IoU}_c=\frac{\mathrm{TP}_c}{\mathrm{TP}_c+\mathrm{FP}_c+\mathrm{FN}_c},\quad
\mathrm{F1}_c=\frac{2\mathrm{TP}_c}{2\mathrm{TP}_c+\mathrm{FP}_c+\mathrm{FN}_c},\\
\mathrm{mIoU}=\frac{1}{|\mathcal{C}|}\sum_{c\in\mathcal{C}}\mathrm{IoU}_c,\quad
\mathrm{F1}=\frac{1}{|\mathcal{C}|}\sum_{c\in\mathcal{C}}\mathrm{F1}_c.
\end{gathered}
\end{equation}
}
Tabs.~\ref{tab:additional_metrics}, \ref{tab:ablation_additional_metrics}, \ref{tab:hyperparameter_additional_metrics}, and \ref{tab:manual_additional_metrics} provide the additional metrics corresponding to Tabs.~\ref{tab:main_results}, \ref{tab:ablation_studies}, \ref{tab:hyperparameter}, and \ref{tab:manual_comparison}, respectively. All three metrics are macro-averaged over the categories in $\mathcal{C}$ and reported as percentages.

\begin{table}[H]
\centering
{\small
\setlength{\tabcolsep}{0.8pt}
\begin{tabular*}{\columnwidth}{@{\extracolsep{\fill}}c c c c c c c@{}}
\toprule
\multicolumn{4}{c}{\textbf{Components}} &
\multirow{2}{*}{\textbf{mAUC (\%)} $\uparrow$} &
\multirow{2}{*}{\textbf{mIoU (\%)} $\uparrow$} &
\multirow{2}{*}{\textbf{F-1 (\%)} $\uparrow$} \\
\cmidrule(r){1-4}
LC & MV & PO & MI & & & \\
\midrule
-- & -- & -- & -- & 98.5 & 89.8 & 94.4 \\
$\checkmark$ & -- & -- & -- & 99.2 & 92.4 & 95.9 \\
$\checkmark$ & $\checkmark$ & -- & -- & 98.9 & 77.4 & 92.8 \\
$\checkmark$ & $\checkmark$ & $\checkmark$ & -- & \textbf{99.4} & 95.1 & 97.5 \\
$\checkmark$ & -- & -- & $\checkmark$ & 99.3 & 94.3 & 97.0 \\
\midrule
$\checkmark$ & $\checkmark$ & $\checkmark$ & $\checkmark$ & 99.1 & \textbf{95.7} & \textbf{97.8} \\
\bottomrule
\end{tabular*}
}
\caption{Additional metrics corresponding to Tab.~\ref{tab:ablation_studies}.}
\label{tab:ablation_additional_metrics}
\end{table}

\begin{table}[H]
\centering
{\small
\setlength{\tabcolsep}{2.0pt}
\begin{tabular*}{\columnwidth}{@{\extracolsep{\fill}}c c c c@{}}
\toprule
$k$ &
\textbf{mAUC (\%)} $\uparrow$ &
\textbf{mIoU (\%)} $\uparrow$ &
\textbf{F-1 (\%)} $\uparrow$ \\
\midrule
0.01 & 99.1 & 93.4 & 96.5 \\
0.05 & \textbf{99.3} & 93.4 & 96.5 \\
0.10 & 99.1 & \textbf{95.7} & \textbf{97.8} \\
0.15 & \textbf{99.3} & 94.6 & 97.2 \\
0.20 & 99.0 & 92.4 & 95.9 \\
\bottomrule
\end{tabular*}
}
\caption{Additional metrics corresponding to Tab.~\ref{tab:hyperparameter}.}
\label{tab:hyperparameter_additional_metrics}
\end{table}

\begin{table}[H]
\centering
{\small
\setlength{\tabcolsep}{2.0pt}
\begin{tabular*}{\columnwidth}{@{\extracolsep{\fill}}c c c c@{}}
\toprule
\textbf{Method} &
\textbf{mAUC (\%)} $\uparrow$ &
\textbf{mIoU (\%)} $\uparrow$ &
\textbf{F-1 (\%)} $\uparrow$ \\
\midrule
GLEAM-0.2 & 98.8 & 90.0 & 94.5 \\
GLEAM-0.6 & 98.9 & 90.1 & 94.6 \\
GLEAM-1.0 & 98.5 & 90.9 & 95.1 \\
GLEAM-1.4 & 99.0 & 90.5 & 94.8 \\
GLEAM-1.8 & 98.9 & 88.4 & 93.5 \\
\midrule
\textbf{Ours} & \textbf{99.3} & \textbf{94.3} & \textbf{97.0} \\
\bottomrule
\end{tabular*}
}
\caption{Additional metrics corresponding to Tab.~\ref{tab:manual_comparison}.}
\label{tab:manual_additional_metrics}
\end{table}

\begin{table*}[t]
\centering
\begin{minipage}[t]{0.45\textwidth}
\vspace{0pt}
\centering
{\small
\setlength{\tabcolsep}{2.5pt}
\begin{tabular}{@{}l c@{}}
\toprule
\textbf{Method} & \textbf{Inference time (ms)} $\downarrow$ \\
\midrule
Uncertainty & 366.30 \\
Zhang et al. & 401.61 \\
RayFronts & 934.58 \\
ActiveSGM & 578.03 \\
GLEAM & 31.56 \\
\midrule
\textbf{Ours} & \textbf{27.39} \\
\bottomrule
\end{tabular}
}
\caption{Per-step inference time (RTX 4090).}
\label{tab:inference_time}
\end{minipage}
\par\medskip
\begin{minipage}[t]{\textwidth}
\vspace{0pt}
\centering
{\small
\setlength{\tabcolsep}{0.6pt}
\renewcommand{\arraystretch}{1.0}
\begin{tabular*}{\linewidth}{@{\extracolsep{\fill}}l c c c c c c c@{}}
\toprule
\multirow{2}{*}{\textbf{Method}} &
\multicolumn{2}{c}{\textbf{CCR (\%)} $\uparrow$} &
\multirow{2}{*}{\textbf{OCR (\%)} $\uparrow$} &
\multirow{2}{*}{\textbf{Var} $\downarrow$} &
\multirow{2}{*}{\textbf{mAUC (\%)} $\uparrow$} &
\multirow{2}{*}{\textbf{mIoU (\%)} $\uparrow$} &
\multirow{2}{*}{\textbf{F-1 (\%)} $\uparrow$} \\
\cmidrule(r){2-3}
& car & building & & & & & \\
\midrule
Uncertainty & 55.9 & 59.5 & 57.7 & \textbf{3.3} & 72.0 & 37.8 & 52.8 \\
Zhang et al. & 50.3 & 79.4 & 64.8 & 212.8 & 71.8 & 34.7 & 49.8 \\
RayFronts & 29.5 & 68.5 & 49.0 & 380.6 & 79.6 & 31.3 & 41.2 \\
ActiveSGM & 46.1 & 69.5 & 57.8 & 136.7 & 70.6 & 32.3 & 47.4 \\
GLEAM & 66.8 & 87.5 & 77.1 & 107.8 & 87.2 & 44.6 & 52.4 \\
\midrule
\textbf{Ours} & \textbf{77.6} & \textbf{89.4} & \textbf{83.5} & 34.9 & \textbf{92.2} & \textbf{47.4} & \textbf{61.9} \\
\bottomrule
\end{tabular*}
}
\caption{UrbanBIS results. Metrics are in \% except Var.}
\label{tab:urbanbis_results}
\end{minipage}
\par\medskip
\begin{minipage}[t]{\textwidth}
\centering
{\small
\setlength{\tabcolsep}{2.0pt}
\begin{tabular*}{\linewidth}{@{\extracolsep{\fill}}c c c c c c c c c@{}}
\toprule
\multirow{2}{*}{\textbf{Configuration}} &
\multicolumn{3}{c}{\textbf{CCR (\%)} $\uparrow$} &
\multirow{2}{*}{\textbf{OCR (\%)} $\uparrow$} &
\multirow{2}{*}{\textbf{Var} $\downarrow$} &
\multirow{2}{*}{\textbf{mAUC (\%)} $\uparrow$} &
\multirow{2}{*}{\textbf{mIoU (\%)} $\uparrow$} &
\multirow{2}{*}{\textbf{F-1 (\%)} $\uparrow$} \\
\cmidrule(r){2-4}
& $c_{\text{small}}$ & $c_{\text{medium}}$ & $c_{\text{large}}$ & & & & & \\
\midrule
LC/LC & 84.3 & \textbf{96.8} & 99.2 & 93.4 & 42.9 & 99.2 & 92.4 & 95.9 \\
LC/LC+MI & 86.5 & 96.3 & \textbf{99.8} & 94.2 & 31.7 & 99.1 & 92.4 & 95.9 \\
LC+MI/LC & 87.3 & 96.5 & 99.6 & 94.5 & 27.6 & 99.0 & 93.3 & 96.4 \\
\textbf{LC+MI/LC+MI} & \textbf{88.3} & 96.2 & 99.7 & \textbf{94.7} & \textbf{22.7} & \textbf{99.3} & \textbf{94.3} & \textbf{97.0} \\
\bottomrule
\end{tabular*}
}
\caption{Post-training module replacement without retraining; X/Y denotes motion/likelihood source checkpoints.}
\label{tab:cross_module}
\end{minipage}
\end{table*}

\subsection{MI Regularization}

To further investigate the impact of mutual information regularization, we conduct two experiments.

First, to isolate the effect of MI regularization on each policy branch, we conduct a $2\times2$ post-training module-swapping experiment, where the motion and likelihood modules are independently selected from the LC and LC+MI trained models. The results are shown in Tab.~\ref{tab:cross_module}. Replacing either LC module with its LC+MI counterpart improves OCR and reduces Var while holding the other module fixed, indicating that the benefit of dependence regularization is encoded in both learned modules rather than arising solely from their joint co-adaptation. Combining both LC+MI modules yields the strongest overall performance. Since all hybrids are evaluated without retraining and the CLUB estimator is absent at inference, these results provide controlled evidence that MI regularization improves the learned motion and likelihood functions themselves.

Second, we run each trained LC and LC+MI model once in the test environments and collect all paired projected motion and calibration features. We arrange the features as $\mathbf{X}_\alpha=[\mathbf{z}_{\alpha,1},\ldots,\mathbf{z}_{\alpha,N}]^\top$ and $\mathbf{X}_\beta=[\mathbf{z}_{\beta,1},\ldots,\mathbf{z}_{\beta,N}]^\top$, and center each feature dimension to obtain $\bar{\mathbf{X}}_\alpha$ and $\bar{\mathbf{X}}_\beta$. Their covariance and cross-covariance matrices are
\begin{equation}
\mathbf{C}_{ij}=\frac{1}{N-1}\bar{\mathbf{X}}_i^\top\bar{\mathbf{X}}_j,
\qquad i,j\in\{\alpha,\beta\}.
\end{equation}
The canonical correlations $\rho_1\geq\cdots\geq\rho_r$ are the singular values of the whitened cross-covariance matrix
\begin{equation}
\mathbf{C}_{\alpha\alpha}^{-\frac{1}{2}}
\mathbf{C}_{\alpha\beta}
\mathbf{C}_{\beta\beta}^{-\frac{1}{2}},
\end{equation}
where $r=\min(d_\alpha,d_\beta)$. With $K=\min(10,r)$, the reported statistics are
\begin{equation}
\begin{aligned}
\mathrm{CCA}_{\max} &= \rho_1, \\
\mathrm{CCA}_{\mathrm{mean}} &= \frac{1}{r}\sum_{i=1}^{r}\rho_i, \\
\mathrm{CCA}_{\mathrm{top\text{-}10}} &= \frac{1}{K}\sum_{i=1}^{K}\rho_i.
\end{aligned}
\end{equation}
The results are reported in Tab.~\ref{tab:feature_cca}. LC+MI reduces Mean CCA by $27.85\%$ and Top-10 CCA by $22.76\%$, showing that MI regularization suppresses overall and dominant shared linear dependence between the two branches. Max CCA remains near one, indicating that the regularizer preserves necessary coupling rather than completely decorrelating the features.

\raggedbottom
\begin{table}[H]
\centering
{\small
\setlength{\tabcolsep}{2.2pt}
\begin{tabular}{@{}lrrrr@{}}
\toprule
\textbf{Metric} & \textbf{LC} & \textbf{LC+MI} & $\boldsymbol{\Delta}$ & \textbf{Rel. $\boldsymbol{\Delta}$} \\
\midrule
Max CCA    & 0.9910 & 0.9938 & $+0.0027$ & $+0.28\%$ \\
Mean CCA   & 0.0625 & 0.0451 & $-0.0174$ & $-27.85\%$ \\
Top-10 CCA & 0.6372 & 0.4922 & $-0.1450$ & $-22.76\%$ \\
\bottomrule
\end{tabular}
}
\caption{Projected-feature CCA with unrounded differences.}
\label{tab:feature_cca}
\end{table}

\begin{figure*}[!t]
    \centering
    \includegraphics[width=0.72\textwidth]{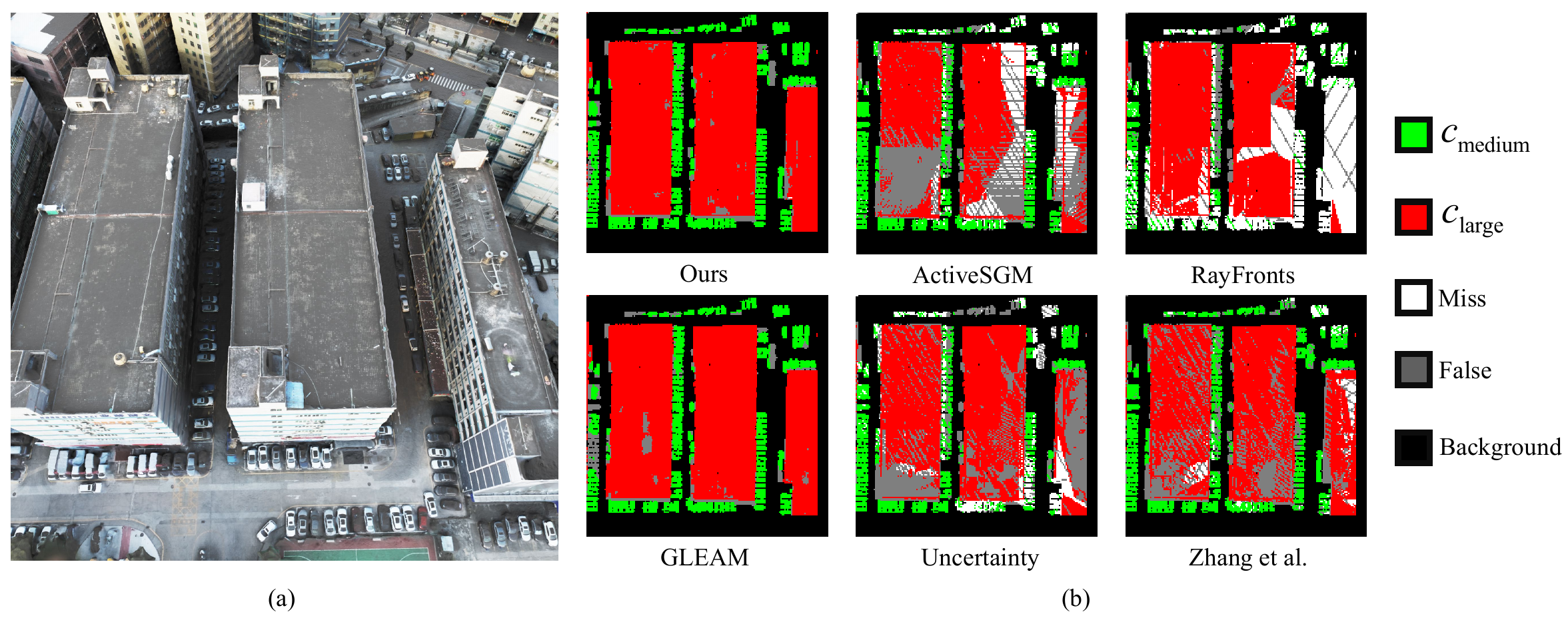}
    \caption{Qualitative results on the real-world UrbanBIS dataset. (a) Visualization in Isaac Sim of the 3D reconstructed mesh of the test scene provided by UrbanBIS. (b) Semantic reconstruction maps produced by different methods.}
    \label{fig:urbanbis}
\end{figure*}

\subsection{Inference Efficiency}

We measure the per-step inference time of all methods on a single RTX 4090 under the same evaluation setup. As reported in Tab.~\ref{tab:inference_time}, our method requires $27.39$ ms per step, which is significantly faster than the planning-based methods. The CLUB estimator is used only during training and is discarded at inference.

\subsection{Evaluation on real-world UrbanBIS data}

All methods are evaluated on UrbanBIS \cite{yang2023urbanbis}. We use the photos of the dataset to reconsturct mesh and use the labeled point cloud to calculate groundtruth semantic map. Training-free baselines are deployed directly; learning-based models transfer are trained using the simulator and directly deployed to the UrbanBIS dataset without fine-tuning. Tab.~\ref{tab:urbanbis_results} and Fig.~\ref{fig:urbanbis} report quantitative and qualitative results. Despite mesh distortions compared to the simulator caused by limited photos, our method performs the best.

\putbib[appendix]
\end{bibunit}

\end{document}